# Approximate Inference and Constrained Optimization

Tom Heskes　　Kees Albers　　Bert Kappen
SNN, University of Nijmegen
Geert Grooteplein 21, 6525 EZ, Nijmegen, The Netherlands

## Abstract

Loopy and generalized belief propagation are popular algorithms for approximate inference in Markov random fields and Bayesian networks. Fixed points of these algorithms correspond to extrema of the Bethe and Kikuchi free energy (Yedidia et al., 2001). However, belief propagation does not always converge, which motivates approaches that explicitly minimize the Kikuchi/Bethe free energy, such as CCCP (Yuille, 2002) and UPS (Teh and Welling, 2002). Here we describe a class of algorithms that solves this typically *non-convex* constrained minimization problem through a sequence of *convex* constrained minimizations of upper bounds on the Kikuchi free energy. Intuitively one would expect tighter bounds to lead to faster algorithms, which is indeed convincingly demonstrated in our simulations. Several ideas are applied to obtain tight convex bounds that yield dramatic speed-ups over CCCP.

## 1 Introduction

Loopy and generalized belief propagation are variational algorithms for approximate inference in Markov random fields and Bayesian networks. Fixed points of loopy and generalized belief propagation have been shown to correspond to extrema of the so-called Bethe and Kikuchi free energy, respectively (Yedidia et al., 2001). However, convergence of loopy and generalized belief propagation to a stable fixed point is not guaranteed and new algorithms have therefore been derived that explicitly minimize the Bethe and Kikuchi free energy (Yuille, 2002; Teh and Welling, 2002). Alas, these algorithms tend to be rather slow and the goal in this article is to come up with faster alternatives.

As we will see in Section 2, minimization of the Kikuchi free energy corresponds to a usually non-convex constrained minimization problem. Non-convex constrained minimization problems are known to be rather difficult, so in Section 3 we will first derive conditions for the Kikuchi free energy to be convex. In Section 4 we will then derive a class of converging double-loop algorithms, in which each inner loop corresponds to constrained minimization of a *convex bound* on the Kikuchi free energy and each outer-loop step to a recalculation of this bound. Based on the intuition that tighter bounds yield faster algorithms, we come up with several ideas to construct tight bounds. The simulations in Section 5 illustrate the use of tight convex bounds. Implications are discussed in Section 6.

## 2 Cluster Variation Method

The exact joint distribution for both undirected (Markov random fields) and directed (Bayesian networks) graphical models can be written in the factorized form

$$P_{\text{exact}}(X) = \frac{1}{Z} \prod_\alpha \Psi_\alpha(X_\alpha) \,, \qquad (1)$$

with $\Psi_\alpha$ potentials, functions defined on the potential subsets $X_\alpha$ and $Z$ the proper normalization constant. Computing this normalization constant, as well as computing marginals on subsets of variables, in principle requires summation over an exponential number of states. To circumvent this exponential summation there are two kinds of approaches: sampling techniques and variational methods.

Variational methods are based on tractable approximations of the Helmholtz free energy

$$F(P) = E(P) - S(P),  \qquad (2)$$

with the energy

$$E(P) \equiv -\sum_\alpha \sum_{X_\alpha} P(X_\alpha) \psi_\alpha(X_\alpha),$$

where $\psi_\alpha(X_\alpha) \equiv \log \Psi_\alpha(X_\alpha)$, and the entropy

$$S(P) \equiv -\sum_X P(X) \log P(X).$$

Functional minimization of $F(P)$ with respect to $P(X)$ under the constraint that $P(X)$ is properly normalized yields $P_{\text{exact}}(X)$. Furthermore, the partition function $Z$ then follows from $-\log Z = F(P_{\text{exact}})$. The variational approximations of the exact free energy (2) can be roughly divided into two classes, the "mean-field" and the "cluster variation" methods. In the cluster variation method (CVM), we represent the probability distribution $P(X)$ through a large number of (possibly overlapping) probability distributions, each describing a subset (cluster) of variables. The minimal choice of these clusters are the subsets $X_\alpha$ that specify the factorization of the potentials. Roughly speaking, the larger the size of the clusters, the more accurate the approximation, but the higher the computational complexity (exponential in the size of the clusters). Without loss of generality, we redefine the subsets $X_\alpha$ in (1) to be the clusters used in the CVM.

Given these "outer clusters" $\alpha$, the so-called Kikuchi approximation of the free energy (2) then leaves the energy term as is and approximates the entropy $S(P) \approx S_{\text{Kik}}(P)$ through a combination of marginal entropies:

$$S_{\text{Kik}}(P) = \sum_\alpha S_\alpha(P) + \sum_\beta c_\beta S_\beta(P), \qquad (3)$$

with

$$S_\alpha(P) \equiv -\sum_{X_\alpha} P(X_\alpha) \log P(X_\alpha)$$

Here the parameters $c_\beta$ are referred to as Moebius or overcounting numbers. The "variable subsets" $x_\beta$, written in lower case to distinguish them from the outer clusters $X_\alpha$, are subsets and typically intersections of two or more outer clusters. In the original CVM (Kikuchi, 1951), the variable subsets consist of all intersections of the outer clusters, intersections of intersections, and so on. With $V$ the collection of all variable subsets and $U$ the collection of all outer clusters, the overcounting numbers in the original CVM follow Moebius formula

$$c_\alpha = 1 \ \forall_{\alpha \in U} \text{ and } c_\gamma = 1 - \sum_{\gamma' \supset \gamma} c_{\gamma'} \ \forall_{\gamma \in V}. \qquad (4)$$

The overcounting numbers for the variable subsets are usually negative, but can also be positive (e.g., for intersections of intersections). We will refer to the respective collections as $V_-$ and $V_+$. The collection $R \equiv U \cup V$ of all "regions" is a so-called partially ordered set or "poset" where the ordering is defined with respect to the inclusion operator $\subset$ (Pakzad and Anantharam, 2002; McEliece and Yildirim, 2003). It can be visualized with a region graph or Hasse diagram (see (Yedidia et al., 2002)). Several extensions, with other constraints on the choice of variable subsets and overcounting numbers, have been proposed recently. An overview can be found in (Yedidia et al., 2002). Here we will call any approximation of the entropy as in (3) a Kikuchi approximation, with the Bethe approximation the special case of non-overlapping variable subsets.

The Kikuchi approximation of the free energy only depends on the marginals $P(X_\alpha)$ and $P(x_\beta)$. We now replace minimization of the free energy over the joint distribution $P(X)$ by minimization of the Kikuchi free energy

$$F_{\text{Kik}}(Q) = \sum_\alpha E_\alpha(Q_\alpha) - S_{\text{Kik}}(\bullet), \qquad (5)$$

with

$$E_\alpha(Q_\alpha) \equiv -\sum_{X_\alpha} Q_\alpha(X_\alpha) \psi_\alpha(X_\alpha),$$

over consistent and normalized pseudo-marginals $Q = \{Q_\alpha, Q_\beta\}$, i.e., under the consistency and normalization constraints

$$\sum_{x_{\gamma'\setminus\gamma}} Q_{\gamma'}(x_{\gamma'}) = Q_\gamma(x_\gamma) \quad \forall_{\gamma'\supset\gamma}$$

$$\sum_{x_\gamma} Q_\gamma(x_\gamma) = 1 \quad \forall_\gamma. \qquad (6)$$

Referring to the class of consistent and normalized pseudo-marginals as $\mathcal{Q}$, we have $-\log Z \approx \min_{Q\in\mathcal{Q}} F_{\text{Kik}}(Q)$. Furthermore, the hope is that the pseudo-marginals $Q_\alpha(X_\alpha)$ corresponding to this minimum are accurate approximations of the exact marginals $P_{\text{exact}}(X_\alpha)$. The Kikuchi free energy and corresponding marginals are exact if the region graph turns out to be singly-connected.

Thus, our task is to minimize the Kikuchi free energy with respect to a set of pseudo-marginals under linear constraints. Constrained minimization is relatively straightforward for convex problems. Therefore, we will first discuss conditions under which the Kikuchi free energy is effectively convex. Then we will consider the more general case of a non-convex Kikuchi free energy.

## 3 Convex Kikuchi Free Energy

In reasoning about convexity, we can disregard the energy term because it is linear in $Q_\alpha$. The entropy terms give either a convex or a concave contribution, depending on whether the corresponding overcounting numbers are positive or negative, respectively. Now, in most if not all relevant cases, there *are* negative overcounting numbers, which makes the Kikuchi free energy (5) non-convex in $\{Q_\alpha, Q_\beta\}$. But perhaps, for example using the constraints to eliminate subset marginals $Q_\beta$ in favor of outer cluster marginals $Q_\alpha$, we can turn the Kikuchi free energy into a functional that *is* convex in $\{Q_\alpha, Q_\beta\}$. Following (Pakzad and Anantharam, 2002), we therefore call a function *"convex over the constraint set"* if, substituting (some of) the constraints, we can turn the possibly non-convex function into a convex one. The idea, formulated in the following theorem, is then that the Kikuchi free energy is convex over the constraint set if we can compensate the concave contributions of the negative variable subsets $\beta \in V_-$ by the convex contributions of outer clusters and positive variable subsets $\gamma \in R_+$ ($R_+ \equiv U \cup V_+$).

**Theorem 3.1** *The Kikuchi free energy is convex over the set of consistency constraints if there exists an "allocation matrix" $A_{\gamma\beta}$ between positive regions $\gamma \in R_+$ and negative variable subsets $\beta \in V_-$ satisfying*

1. $\gamma$ *can be used to compensate for $\beta$:*

   $A_{\gamma\beta} \neq 0$ *only if* $\gamma \supset \beta$

2. *positive compensation:* $A_{\gamma\beta} \geq 0$

3. *sufficient resources:*

   $$\sum_{\beta\subset\gamma} A_{\gamma\beta} \leq c_\gamma \quad \forall_{\gamma\in R_+}$$

4. *sufficient compensation:*

   $$\sum_{\gamma\supset\beta} A_{\gamma\beta} \geq |c_\beta| \quad \forall_{\beta\in V_-}$$

**Sketch of proof.** The combination of a convex entropy contribution from $\gamma \in R_+$ with the concave entropy contribution from $\beta \in V_-$, where $\beta \subset \gamma$,

$$-S_\gamma(Q_\gamma) + S_\beta(Q_\beta)$$

is convex over the constraint $Q_\gamma(x_\beta) = Q_\beta(x_\beta)$. The proof then follows from a decomposition of the entropy into terms that are all convex when the conditions are satisfied. ∎

The conditions of Theorem 3.1 can be checked with a linear program. It follows from this theorem that the Bethe free energy is convex over the constraint set if the graph contains a single loop. Furthermore, if the graph contains two or more connected cycles, the conditions fail. A similar theorem with the same corollary is given in (Pakzad and Anantharam, 2002; McEliece and Yildirim, 2003).

If the Kikuchi free energy is convex over the constraint set, it must have a unique minimum. The message passing algorithm outlined in Algorithm 1 then converges to this minimum, with perhaps a little damping in the case of negative $c_\beta$ (see similar argumentation in (Wainwright et al., 2003); $c_\beta = 0$ is just fine). Algorithm 1 is a specific instance

---

**Algorithm 1** Message-passing algorithm.
1: **while** ¬converged **do**
2:   **for all** $\beta \in V$ **do**
3:     **for all** $\alpha \in U, \alpha \supset \beta$ **do**
4:       $Q_\alpha(x_\beta) = \sum_{X_{\alpha\setminus\beta}} Q_\alpha(X_\alpha)$
5:       $\mu_{\alpha\to\beta}(x_\beta) = \dfrac{Q_\alpha(x_\beta)}{\mu_{\beta\to\alpha}(x_\beta)}$
6:     **end for**
7:     $Q_\beta(x_\beta) \propto \prod_{\alpha\supset\beta} \mu_{\alpha\to\beta}(x_\beta)^{\frac{1}{n_\beta+c_\beta}}$
8:     **for all** $\alpha \in U, \alpha \supset \beta$ **do**
9:       $\mu_{\beta\to\alpha}(x_\beta) = \dfrac{Q_\beta(x_\beta)}{\mu_{\alpha\to\beta}(x_\beta)}$
10:      $Q_\alpha(X_\alpha) \propto \Psi_\alpha(X_\alpha) \prod_{\beta\subset\alpha} \mu_{\beta\to\alpha}(x_\beta)$
11:   **end for**
12:  **end for**
13: **end while**

---

of generalized belief propagation and reduces to standard loopy belief propagation for the Bethe free energy, where $c_\beta = 1 - n_\beta$ with $n_\beta$ the number of neighboring outer clusters.

## 4 Double-Loop Algorithms

### 4.1 General Procedure

Fixed points of Algorithm 1 correspond to extrema of the Kikuchi free energy under the appropriate constraints (Yedidia et al., 2001). However, in practice this single-loop algorithm does not always converge and we have to resort to double-loop algorithms to guarantee convergence to a minimum of the Kikuchi free energy. Here we will introduce a class of such algorithms based on the following theorem.

**Theorem 4.1** *Given an at least twice differentiable function $F_{\text{conv}}(Q, Q')$ with properties*

1. $F_{\text{conv}}(Q, Q') \geq F_{\text{Kik}}(Q) \quad \forall_{Q,Q'\in\mathcal{Q}}$
2. $F_{\text{conv}}(Q, Q) = F_{\text{Kik}}(Q) \quad \forall_{Q\in\mathcal{Q}}$
3. $F_{\text{conv}}(Q, Q')$ *is convex in* $Q \in \mathcal{Q} \quad \forall_{Q'\in\mathcal{Q}}$

*the algorithm*

$$Q_{n+1} = \underset{Q\in\mathcal{Q}}{\operatorname{argmin}} F_{\text{conv}}(Q, Q_n), \qquad (7)$$

*with $Q_n$ the pseudo-marginals at iteration $n$, is guaranteed to converge to a local minimum of the Kikuchi free energy $F_{\text{Kik}}(Q)$ under the appropriate constraints.*

**Proof.** It is immediate that the Kikuchi free energy decreases with each iteration:

$$\begin{aligned} F_{\text{Kik}}(Q_{n+1}) &\leq F_{\text{conv}}(Q_{n+1}, Q_n) \\ &\leq F_{\text{conv}}(Q_n, Q_n) = F_{\text{Kik}}(Q_n), \end{aligned}$$

where the first inequality follows from condition 1 (upper bound) and the second from the definition of the algorithm. Condition 2 (touching) in combination with differentiability ensures that the algorithm is only stationary in points where the gradient of $F_{\text{Kik}}$ is zero. By construction $Q_n \in \mathcal{Q}$ for all $n$. ∎

Convexity of $F_{\text{conv}}$ has not been used to establish the proof. However, constrained minimization of a convex functional is much simpler than constrained minimization of a nonconvex functional. This general idea, replacing the minimization of a complex functional by the consecutive minimization of easier to handle upper bounds, forms the basis of popular algorithms such as the EM algorithm (Neal and Hinton, 1998), iterative scaling/iterative proportional fitting (Jiroušek and Přeučil, 1995), and algorithms for non-negative matrix factorization (Lee and Seung, 2001). Intuitively, the tighter the bound, the faster the algorithm.

### 4.2 Bounding the Concave Terms

As a first step, to lay out the main ideas, we build a convex bound by removing all concave entropy contributions for $\beta \in V_-$. To do so, we will make use of the linear bound

$$S_\beta(Q_\beta) = -\sum_{x_\beta} Q_\beta(x_\beta) \log Q_\beta(x_\beta) \leq$$
$$-\sum_{x_\beta} Q_\beta(x_\beta) \log Q'_\beta(x_\beta) \equiv S_\beta(Q_\beta, Q'_\beta), \quad (8)$$

which directly follows from $\text{KL}(Q_\beta, Q'_\beta) \geq 0$ with KL the Kullback-Leibler divergence. Our choice $F_{\text{conv}}$ then reads

$$F_{\text{conv1}}(Q, Q') = \sum_\alpha E_\alpha(Q_\alpha) - \sum_\alpha S_\alpha(Q_\alpha)$$
$$- \sum_{\beta\in V_+} c_\beta S_\beta(Q_\beta) - \sum_{\beta\in V_-} c_\beta S_\beta(Q_\beta, Q'_\beta).$$

It is easy to check that this functional satisfies all conditions for Theorem 4.2.

Next we make the pleasant observation that, using the constraints (6) and for fixed $Q'$, we can rewrite $F_{\text{conv1}}$ in the "normal form" (5) through a redefinition of the overcounting numbers and the potentials. The new overcounting numbers $\tilde{c}_\beta$ refer to all *unbounded* entropy contributions; for $F_{\text{conv1}}$

$$\tilde{c}_\beta = 0 \;\; \forall_{\beta \in V_-} \text{ and } \tilde{c}_\beta = c_\beta \;\; \forall_{\beta \in V_+} . \qquad (9)$$

The *bounded* entropy contributions can be incorporated in the energy term by redefining the (log) potentials, for example through

$$\tilde{\psi}_\alpha(X_\alpha) = \psi_\alpha(X_\alpha) - \sum_{\beta \subset \alpha} \frac{(c_\beta - \tilde{c}_\beta)}{n_\beta} \log Q'_\beta(x_\beta) . \qquad (10)$$

With $F_{\text{conv1}}$ both convex and in normal form, we can use Algorithm 1, with substitutions

$$c_\beta \Leftarrow \tilde{c}_\beta \text{ and } \psi_\alpha \Leftarrow \tilde{\psi}_\alpha , \qquad (11)$$

to solve the constrained problem (7).

The general setting of the double-loop algorithm is as follows.

***Beforehand:*** choose $\tilde{c}_\beta$, e.g. as in (9).

***Outer:*** compute $\tilde{\psi}_\alpha$ from (10) with $Q' = Q_n$.

***Inner:*** Algorithm 1 with (11) yielding $Q_{n+1}$.

### 4.3 Bounding Convex Terms As Well

In many cases we can make the algorithm both better and simpler by bounding not only the concave, but also the convex entropy contributions. That is, we define $F_{\text{conv2}}$ by setting

$$\tilde{c}_\beta = 0 \;\; \forall_{\beta \in V} . \qquad (12)$$

The basic algorithm and potential updates (10) stay the same, but now with (12) instead of (9).

The algorithm based on $F_{\text{conv2}}$ is simpler than the one based on $F_{\text{conv1}}$ because it typically runs over less variable subsets: all variable subsets that have zero overcounting number and are not direct intersections of outer clusters can be left out in the inner loop.

From (8), but now applied to the positive variable subsets, it is clear that $F_{\text{conv2}}(Q, Q') \leq F_{\text{conv1}}(Q, Q')$: when it is a bound, $F_{\text{conv2}}$ is a tighter bound than $F_{\text{conv1}}$ and we can expect the algorithm based on $F_{\text{conv2}}$ to perform better. It remains to be shown under which conditions $F_{\text{Kik}}(Q) \leq F_{\text{conv2}}(Q, Q')$. This is where the following theorem comes in.

**Theorem 4.2** $F_{\text{conv2}}$ *defined from (12) is a convex bound of the Kikuchi free energy (5) if there exists an "allocation matrix" $A_{\gamma\beta}$ between negative variable subsets $\gamma \in V_-$ and positive variable subsets $\beta \in V_+$ satisfying*

1. $A_{\gamma\beta} \neq 0$ only if $\gamma \supset \beta$
2. $A_{\gamma\beta} \geq 0$
3. $\sum_{\beta \subset \gamma} A_{\gamma\beta} \leq |c_\gamma| \;\; \forall_{\gamma \in V_-}$ (13)
4. $\sum_{\gamma \supset \beta} A_{\gamma\beta} \geq c_\beta \;\; \forall_{\beta \in V_+}$

**Sketch of proof.** We follow the same line of reasoning as the proof of Theorem 3.1. First we note that, if $\beta \subset \gamma$,

$$S_\gamma(Q_\gamma) - S_\beta(Q_\beta) \leq S_\gamma(Q_\gamma, Q'_\gamma) - S_\beta(Q_\beta, Q'_\beta),$$

i.e., if we bound a concave $S_\gamma(Q_\gamma)$, we can incorporate a convex $-S_\beta(Q_\beta)$ to make the bound tighter. Shielding all convex contributions with concave contributions is then again a matter of resource allocation. ∎

As above, the conditions for Theorem 4.2 can be checked with a linear program. In practice, these conditions hold more often than not.

### 4.4 Just Convex over the Constraints

The bounds $F_{\text{conv1}}$ and $F_{\text{conv2}}$ are convex without reference to the constraints. We can make the bound tighter by bounding less concave entropy contributions, but just enough to make it *convex over the constraint set* instead of convex per se. And again, following the ideas in the previous section, we can try to incorporate convex entropy contributions in the concave terms that have to be bounded anyway. This is implemented in the following procedure.

1. Choose $\tilde{c}_\beta \geq c_\beta$ for $\beta \in V_-$ such that

$$-\left\{\sum_\alpha S_\alpha + \sum_{\beta \in V_-} \tilde{c}_\beta S_\beta + \sum_{\gamma \in V_+} c_\gamma S_\gamma\right\}$$

is (just) convex over the constraint set. The remaining $(c_\beta - \tilde{c}_\beta)S_\beta$ will be bounded.

2. With $A$ the corresponding allocation matrix of Theorem 3.1, define the "used resources"

$$\forall_{\gamma \in V_+} \quad \hat{c}_\gamma \equiv \sum_{\beta \in V_-} A_{\gamma\beta}|\tilde{c}_\beta| \leq c_\gamma,$$

and thus "unused resources" $c_\gamma - \hat{c}_\gamma$.

3. To make the bound tighter, incorporate as many of the unused $-(c_\gamma - \hat{c}_\gamma)S_\gamma$ convex contributions as possible in the $(c_\beta - \tilde{c}_\beta)S_\beta$ concave contributions that have to be bounded anyway. Call the corresponding overcounting numbers $c_\gamma - \tilde{c}_\gamma \leq c_\gamma - \hat{c}_\gamma$.

The inner-loop overcounting numbers $\tilde{c}_\beta$ in the first step and $\tilde{c}_\gamma$ in the third can be found with a linear program and again fully specify the convex bound, referred to as $F_{\text{conv3}}$, and the corresponding double-loop algorithm.

### 4.5 Related Work

Although originally formulated in a different way, the CCCP (concave-convex procedure) algorithm of (Yuille, 2002) can also be understood as a particular case of the general procedure outlined in Theorem 4.1. More specifically, it is based on bounding the concave contributions with

$$|c_\beta|S_\beta(Q_\beta) \leq -S_\beta(Q_\beta) + (|c_\beta|+1)S_\beta(Q_\beta, Q'_\beta),$$

which is to be compared with (8). That is, before bounding the concave entropy contributions, part of them are taken over to the "convex side". In terms of the inner-loop overcounting numbers $\tilde{c}_\beta$ this amounts to

$$\tilde{c}_\beta = 1 \quad \forall_{\beta \in V_-} \text{ and } \tilde{c}_\beta = c_\beta \quad \forall_{\beta \in V_+}. \quad (14)$$

This makes the bound less tight[1].

---

[1] In (Yuille, 2002) it is further suggested to take convex terms to the concave side, in particular to set $\tilde{c}_\beta = \max_{\beta'} c_{\beta'} \forall_{\beta \in R}$. This tends to make the bound a lot looser. Here we will stick to the more favorable interpretation based on (14).

The UPS (unified propagation and scaling) algorithm of (Teh and Welling, 2002) also replaces constrained minimization of a nonconvex function by sequential minimization of functions that are convex over the constraint set, fairly similar to our algorithm based on $F_{\text{conv3}}$. The main difference is that where we *bound* part of the concave entropy contributions, UPS *clamps* some of them. This makes UPS considerably less flexible.

In (Wainwright et al., 2003) convex bounds on the exact Helmholtz free energy (2) are presented. In these bounds, the overcounting numbers for the variable subsets still follow the Moebius relationship (4), but the overcounting numbers for the outer clusters are smaller than or equal to 1. Constrained minimization of this bound is very similar to constrained minimization of $F_{\text{conv3}}$ and the algorithm proposed in (Wainwright et al., 2003) is indeed closely related to Algorithm 1.

## 5 Simulations

We have done simulations on quite a number of different problems and problem instances, involving both Markov random fields and Bayesian networks. In Figure 1 we give a few examples, meant to illustrate the general picture that we will summarize below. In our setup, the different algorithms only differ in the (tightness of the) convex bounds used in the inner loop, represented through the inner-loop overcounting numbers $\tilde{c}_\beta$: just convex over the set of constraints as explained in Section 4.4 (solid lines), with all entropy contributions bounded using (12) in Section 4.3 (dotted), with only the concave contributions bounded using (9) in Section 4.2 (dashed), and our rather favorable interpretation (14) of the bound implicit in the CCCP algorithm (dash-dotted). In all cases, the convex constrained minimization in the inner loop is solved by Algorithm 1, which is run until a preset criterion is met (here until the variable subset marginals change less than $10^{-4}$). We report on the Kullback-Leibler divergence between approximate and exact single-node marginals (top row). Where we expect the algorithm based on the tightest bound to converge the fastest in terms of outer-loop iterations, we might need more inner-loop iterations to achieve convergence in the inner loop. Therefore we also plot the number of inner-loop iterations required to meet the convergence criterion (middle row). To make them comparable, the outer-loop iterations on the x-axis are scaled relative to those required for the just-convex algorithm to reach the same level of accuracy. The inner-loop overcounting numbers give an indication of the tightness of the bounds (bottom row: the lower, the tighter), with those for the Kikuchi free energy itself on the first line.

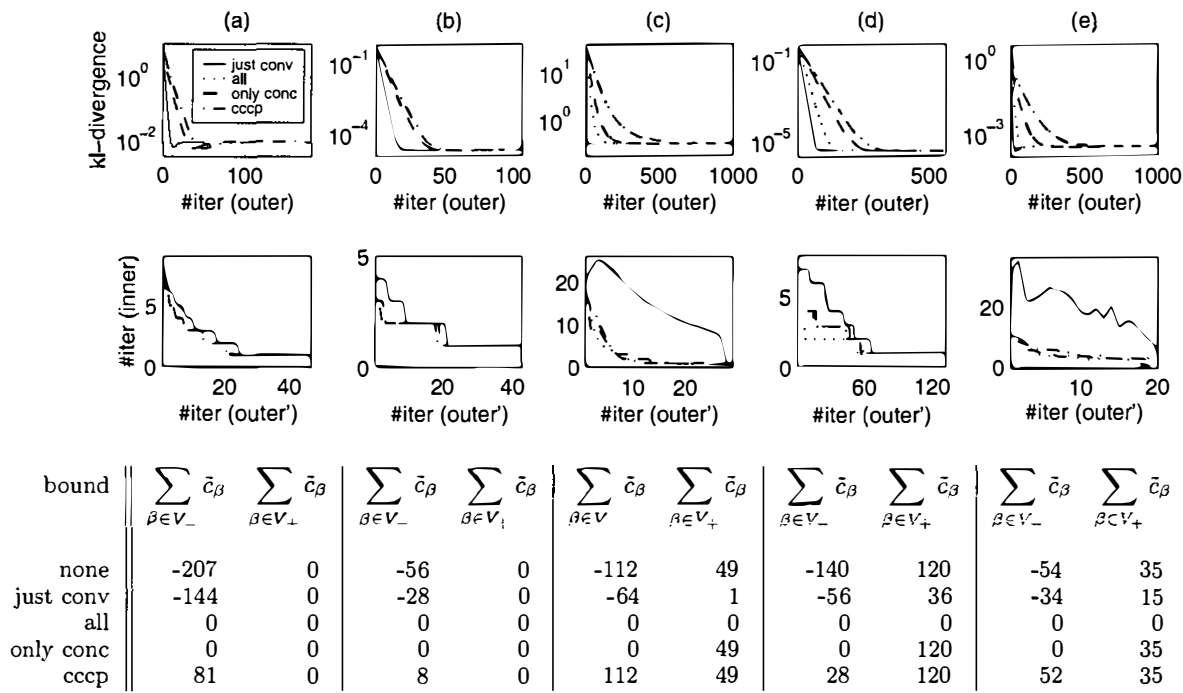

Figure 1: Top: KL-divergence between exact and approximate marginals. Middle: number of inner-loop iterations required to meet a fixed convergence criterion. Bottom: inner-loop overcounting numbers. (a) Bethe and (c) Kikuchi (outer clusters 4 neighbors) on a 9x9 Boltzmann grid. (b) Bethe and (d) Kikuchi (outer clusters all triplets) on an 8-node fully connected Boltzmann machine. (e) Kikuchi on a 20x10 QMR network. See the text for further explanation.

Here we summarize our main experimental findings, based on the simulations visualized in Figure 1 and many other problem instances.

- The tighter the (convex) bound used in the inner loop, the faster the convergence in terms of outer-loop iterations: the ordering in Figure 1 is always (from fastest to slowest) just convex, all bounded, concave bounded, CCCP.

- The number of inner-loop iterations needed to meet a preset convergence criterion sometimes decreases with a looser bound, but never enough to compensate for the slower convergence in the outer loop. For example, in Figure 1(e) the just-convex algorithm uses much more inner-loop iterations per outer-loop iteration than the other three algorithms, but this is compensated by the more than ten-fold speed-up in the outer loop. Note further that the inner-loop convergence criterion is rather strict: all algorithms would probably do just fine with a (much) looser criterion.

- In terms of floating point operations, a looser bound that sets all overcounting numbers in the inner loop to zero, occasionally beats a tighter bound with negative overcounting numbers: the slower convergence in terms of outer-loop iterations is compensated by a more efficient inner loop (see Section 4.3).

## 6 Discussion

This article is based on the perspective that we are interested in *minima* of the Kikuchi free energy under appropriate constraints. Finding such a minimum then becomes a possibly non-convex constrained minimization prob-lem. Here, as well as in other studies, the approach has been to solve this non-convex problem through sequential constrained minimization of convex bounds on the Kikuchi free energy. On the presumption that tighter bounds yield faster algorithms, we have worked out several ideas to construct tight convex bounds. The simulation results clearly validate this presumption and show that the speed-ups can be very significant.

It has been suggested that if generalized/loopy belief propagation does not converge, it makes no sense to explicitly minimize the Kikuchi/Bethe free energy. Others have reported acceptable approximations that a single-loop approach did not manage to converge to (the results in Figure 1(c), (d), and (e) are examples hereof). It seems that there is a definite "middle range" in which generalized/loopy belief propagation does not converge, yet the minimum of the (non-convex) Kikuchi/Bethe free energy does correspond to a fairly accurate approximation of the minimum of the exact Helmholtz free energy.

For convergence of (a damped version of) the single-loop algorithm 1, it is sufficient but not necessary for the bound $F_{conv}$ to be convex over the constraint set. That is, one might well try to start with a tighter non-convex bound, check whether Algorithm 1 converges to a solution that satisfies the constraints and corresponds to a lower Kikuchi free energy, and restart with a looser bound if not. Or even better, perhaps we could come up with conditions, looser than those for Theorem 3.1, based on which we can check beforehand whether Algorithm 1 will converge. These conditions then should take into account not only properties of the (region) graph, but also (the size of) the potentials, perhaps similar to those in (Tatikonda and Jordan, 2002).

## Acknowledgements

This work has been supported in part by the Dutch Technology Foundation STW.